\documentclass[conference]{IEEEtran}

\makeatletter
\def\ps@IEEEtitlepagestyle{%
  \def\@oddfoot{\mycopyrightnotice}%
  \def\@oddhead{\hbox{}\@IEEEheaderstyle\leftmark\hfil}\relax
  \def\@evenhead{\@IEEEheaderstyle\hfil\leftmark\hbox{}}\relax
  \def\@evenfoot{}%
}
\def\mycopyrightnotice{%
  \begin{minipage}{\textwidth}
  \centering \scriptsize
  Copyright~\copyright~2023 IEEE. Personal use of this material is permitted. Permission from IEEE must be obtained for all other uses, in any current or future media, including reprinting/republishing this material for advertising or promotional purposes, creating new collective works, for resale or redistribution to servers or lists, or reuse of any copyrighted component of this work in other works.
  \end{minipage}
}
\makeatother

\IEEEoverridecommandlockouts
\usepackage{cite}
\usepackage{amsmath,amssymb,amsfonts}
\usepackage{algorithmic}
\usepackage{graphicx}
\usepackage{textcomp}
\usepackage{xcolor}

\def\BibTeX{{\rm B\kern-.05em{\sc i\kern-.025em b}\kern-.08em
    T\kern-.1667em\lower.7ex\hbox{E}\kern-.125emX}}
\begin{document}
\thispagestyle{IEEEtitlepagestyle}
\title{FLASH-RL: \underline{F}ederated \underline{L}earning \underline{A}ddressing \underline{S}ystem and \underline{S}tatic \underline{H}eterogeneity using \underline{R}einforcement \underline{L}earning}

\author{\IEEEauthorblockN{Sofiane Bouaziz\IEEEauthorrefmark{1}, Hadjer Benmeziane\IEEEauthorrefmark{2}, Youcef Imine\IEEEauthorrefmark{2}, Leila Hamdad\IEEEauthorrefmark{1}, Smail Niar\IEEEauthorrefmark{2}, Hamza Ouarnoughi\IEEEauthorrefmark{2}}
\IEEEauthorblockA{\IEEEauthorrefmark{1} École nationale Supérieure d’Informatique, Algiers, Algeria }
\IEEEauthorblockA{\IEEEauthorrefmark{2}Univ. Polytechnique Hauts-de-France, CNRS, UMR 8201 - LAMIH, F-59313 Valenciennes, France}}

\maketitle

\begin{abstract}
Federated Learning (FL) has emerged as a promising Machine Learning paradigm, enabling multiple users to collaboratively train a shared model while preserving their local data. 
To minimize computing and communication costs associated with parameter transfer, it is common practice in FL to select a subset of clients in each training round. This selection must consider both system and static heterogeneity. 
Therefore, we
propose FLASH-RL, a framework that utilizes Double Deep Q-Learning (DDQL) to address both system and static heterogeneity in FL. 
FLASH-RL introduces a new reputation-based utility function to evaluate client contributions based on their current and past performances. Additionally, an adapted DDQL algorithm is proposed to expedite the learning process. Experimental results on MNIST and CIFAR-10 datasets have shown FLASH-RL’s effectiveness in achieving a balanced trade-off between model performance and end-to-end latency against existing solutions. Indeed, FLASH-RL reduces latency by up to 24.83\% compared to FedAVG and 24.67\% compared to FAVOR. It also reduces the training rounds by up to 60.44\% compared to FedAVG and +76\% compared to FAVOR. In fall detection using the MobiAct dataset, FLASH-RL outperforms FedAVG by up to 2.82\% in model's performance and reduces latency by up to 34.75\%. Additionally, FLASH-RL achieves the target performance faster, with up to a 45.32\% reduction in training rounds compared to FedAVG.
\end{abstract}

\begin{IEEEkeywords}
Federated Learning, Reinforcement Learning, System heterogeneity, Static heterogeneity, Client selection.
\end{IEEEkeywords}

\section{Introduction}
\label{sec:introduction}
Federated learning (FL) is a promising approach to machine learning (ML) that allows privacy-preserving model training in heterogeneous and distributed networks. 
The motivation behind FL is to make more data available to improve day-to-day lives. 
It enables edge devices to collaborate in order to train a shared ML model while maintaining control of data on the edge level. 
In the context of smart homes, FL offers significant advantages where 
the smart homes are equipped with various sensors that generate a wealth of data about occupants' activities and living environment. 
By applying FL in smart homes~\cite{nguyen_federated_2022}, it becomes possible to develop personalized and context-aware services while respecting privacy.


In FL, each participant (\textit{client's device}) independently performs model training on its local data and communicates the updated model parameters to a central server for aggregation. 
Existing works~\cite{sannara_federated_2021, saloni_anonymised_2021, qin_multi_2021} mainly focus on optimizing this aggregation to improve the performance of the global model. 
In real-world scenarios such as smart homes, the number of clients is large, making it necessary to select the appropriate subset of clients participating in each training round. 
The client selection process plays a crucial role in FL, as it impacts the convergence time, communication overhead, and overall performance of the trained model. 
The shared model is then deployed at the end of the training iterations to allow better performance on each client's device.

A key challenge in FL is the presence of \textit{system} and \textit{static} heterogeneity across the participating clients. 
\textit{System heterogeneity} refers to variations in the computational capabilities and resources of the clients, such as processing power, memory, and network connectivity. 
For instance, devices with limited processing power or unstable network connections may experience delays in completing their local training, leading to overall slower convergence of the global model. 
\textit{Static heterogeneity} pertains to the diversity in the number of samples and the data characteristics across clients, including demographics, classes or labels of interest, data distributions, and sensor types. 
This diversity can have a significant impact on the performance and generalization of the trained model.

Several works~\cite{fu_client_2022} have focused on performing client selection to mitigate the effects of system and static heterogeneity. 
They focus on selecting the best subset of clients that can represent the diversity of the entire client population while also considering their computational capabilities. 
However, these works typically address either system or static heterogeneity separately, lacking simultaneous handling of both aspects.


This paper introduces a novel and efficient FL client selection strategy using reinforcement learning (RL). 
Our approach addresses the challenges of system and static heterogeneity by considering the computational capabilities of clients, such as processing power and network connectivity, along with their data characteristics.

\noindent The contributions presented in this paper are as follows:
 
\begin{itemize}
    \item We define a reputation-based utility function that assesses the performance of each client's local model during training and the latency incurred in training its model with its own data. 
    This function serves as the basis for the reward function, which is optimized using our RL agent. 
    Incorporating the reputation improves the quality and reliability of the selected clients, striking a favorable balance between maximizing the global model performance and minimizing the overall training latency.
    \item We design double deep Q-learning (DDQL) and multi-action selection-based RL algorithms to perform the client's selection during the FL process.
    \item We apply our approach in a real-world use-case which is fall detection in smart homes.  
\end{itemize}


The rest of the paper is organized as follows. Section~\ref{sec:Related_works} provides an overview of the FL process and discusses related works. Section~\ref{sec:Systemheterogneity} explains how latency is computed in FL training rounds. In Section~\ref{sec:Staticheterogneity}, we define the reputation-based utility function for scoring the client's local performance. 
We then give an overview of the client's selection process and describe our RL agent in Section~\ref{sec:Methodology}
Finally, Section~\ref{sec:Experiments} outlines the experimental methodology and presents the results.

\section{Background \& Related Works}
\label{sec:Related_works}
This section provides a background on FL and  
 the heterogeneity problem. A literature on RL-based approaches and the fall detection use-case review is presented as well.   

\subsection{Federated Learning Problem Formulation}

\smallskip
\noindent In FL~\cite{mcmahan_communication-efficient_2017}, we assume the existence of $N$ clients $\left\{P_i\right\}_{i=1}^N$ with their own datasets $\left\{D_i\right\}_{i=1}^N$. 
The aim of FL is to train a global model without having to expose local client data. Thus, for each client $k$, a local ML model is trained to minimize its objective function, denoted $F_k$, and is defined in \ref{eq:fonctionobjclocal}.

\begin{equation}
\label{eq:fonctionobjclocal}
F_k(w)=\frac{1}{n_K} \sum_{i \in D_k} l\left(x_i, y_i ; w_k\right)
\end{equation}

\noindent  where $n_k = |D_k|$ represents the number of training data provided by client $k$, and $l\left(x_i, y_i ; w_k\right)$ corresponds to the loss function of the prediction on the data sample $(x_i, y_i)$ using the local model parameters of client $k$, denoted $w_k$. 
The FL optimization problem is described as follows in Equation \ref{eq:fl_optim}.
\begin{equation}
\label{eq:fl_optim}
\min _{w \in R^d} 	\phi(w)=\sum_{k=1}^N p_k F_k(w) 
\end{equation}

\noindent where $p_k>0$ and $\sum_{i=1}^N p_k=1$.

\smallskip
\noindent At each communication round, the server sends its model parameters to the clients, who in turn train their local models. The clients then send their local parameters back to the server for aggregation. Nevertheless, if the server selects all the clients at the same time, it overloads the network and incurs excessive computation and communication costs. 
Thus, only a subset of the clients are selected to participate in the training. 

\subsection{Static and System Heterogeneity}
\noindent
The client selection needs to account for FL heterogeneity problems~\cite{li_federated_2020}:

\begin{itemize} 
\item[•] \textbf{Static heterogeneity}: Client data is often distributed non-identically and non-independently (non-IID), resulting in heterogeneous data distribution and unbalanced learning between clients. 
For example, countries in the southern hemisphere may have higher rates of skin cancer patients than their counterparts in the northern hemisphere, leading to variations in disease distribution between them.
\item[•] \textbf{System heterogeneity}: The storage, computing, and communication capacities of each client may differ due to the variability of hardware, and network connectivity.
This leads to variability in the transfer time of client parameters. To tackle this issue, \textit{latency} has garnered significant attention within the FL domain, resulting in several works targeting its minimization~\cite{liu_time_2022, li_latency_2022}.
\end{itemize}

\noindent 
Even though existing FL algorithms~\cite{mcmahan_communication-efficient_2017, li_federated_2020-1} achieve good performances, they remain unstable in highly heterogeneous environments. 

\noindent In \cite{moming_astrae_2019}, they proposed Astraeaa, a self-balancing FL framework to mitigate imbalances in data samples. However, it only addresses one type of static heterogeneity and does not account for system heterogeneity. In contrast,\cite{lian_fednorm_2022} introduced FedNorm as a framework that handles both static and system heterogeneity in FL. Although, their approach remains static and lacks the incorporation of intelligent algorithms.




\subsection{Reinforcement Learning-based Solutions}




\smallskip
\noindent The use of RL for client selection in FL has shown great potential in recent works~\cite{wang_optimizing_2020, zhang_adaptive_2021, rjoub_trust-augmented_2022, zhang_multi-agent_2022}. 
The latter proposes an optimization framework that enables adaptive client selection by considering various factors, such as data distribution and hardware resources. 
This enables the client selection strategy to adapt to changes in 
federated environment, improving the performance and efficiency of the global model.

\smallskip

\noindent Table \ref{tab:FLalgocomparative} presents a comparative study of RL-based client selection approaches organized by the used RL algorithm, their ability to incorporate static and system heterogeneity, and the number of RL agents involved.

\begin{table}[ht]
\centering
\caption{Comparison of state-of-the-art RL client selection.}\vspace{-0.5cm}
\begin{tabular}[t]{lcccc}
\hline
Approach & Algorithm & Static & System & Number \\
         &           & heterogeneity & heterogeneity & of agents \\
\hline
~\cite{wang_optimizing_2020} & DDQL & \checkmark & x & $1$ \\
~\cite{zhang_adaptive_2021} & DDQL & x & \checkmark & $1$ \\
~\cite{rjoub_trust-augmented_2022} & DDQL & x & \checkmark  & $1$ \\
~\cite{zhang_multi-agent_2022} & MARL & \checkmark & \checkmark & $N$ \\
\hline
\label{tab:FLalgocomparative}
\end{tabular}
\vspace{-0.5cm}
\end{table}

\noindent In~\cite{wang_optimizing_2020}, the authors introduce FAVOR, an experience-based control framework that leverages the DDQL algorithm for client selection considering static heterogeneity. However, the training of the DDQL model relies on a single client, which may impede rapid convergence of the agent. 
Similarly,~\cite{zhang_adaptive_2021} and~\cite{rjoub_trust-augmented_2022} use DDQL in client selection, with a focus on addressing system heterogeneity. 
Notably, the distinctive aspect of ~\cite{zhang_adaptive_2021} lies in its consideration of a variable number of clients. In~\cite{zhang_multi-agent_2022}, a multi-agent RL (MARL) system was introduced to address both system and static heterogeneity in client selection. 
This approach employed a one-to-one correspondence between the number of agents in the system and the number of clients. 
However, using multiple agents results in additional resource demands, making it a resource-intensive solution.


\subsection{Fall Detection Use-case}

\smallskip
\noindent Fall detection is a classification method used to distinguish between regular human activities and instances of falling. Temporal data captured by wearable sensors, including accelerometer, gyroscope, and magnetometer, are used to detect falls using either threshold-based or ML-based methods. 
The wearable sensors are commonly deployed in smart home environments. 
However, accessing the data generated by them can pose confidentiality challenges. 

\smallskip
\noindent In centralized solutions, ~\cite{Wayan_fall_2019} proposes to use Long Short-Term Memory neural network to develop an algorithm for distinguishing Activities of Daily Living (ADL). In~\cite{diana_fall_2019}, the authors proposed the application of ensemble methods to classify temporal data from accelerometers as ADL or fall.


\smallskip
\noindent FL-based solutions feature the work of~\cite{wu_fedhome_2020} that presents \textit{FedHome}, a cloud-based FL framework designed for personalized in-home health monitoring. 
Their approach primarily focuses on model size reduction and 
enhancing performances across clients. 
Our work focuses on optimizing the client selection process using an RL-based methodology. 

\section{Proposed Approach}
In this section, we describe the objective functions employed to account for system and static heterogeneity. Subsequently, we propose a DDQL-based solution that incorporates them into the reward function to determine the best clients to select during each training round.  

\subsection{Handling System Heterogeneity: Latency}
\label{sec:Systemheterogneity}
It is crucial to account for the varying computational capabilities and connectivity conditions among the clients. In this work, we consider latency as the objective function to handle system heterogeneity. Minimizing latency involves optimizing the time it takes for data to be transmitted, processed, and returned to the clients during the FL process.

\smallskip

\noindent 

\noindent We define the latency of a selected client $k$ during training round $t$ as the sum of its local computing time, denoted as $T_t^{\text {local}_k}$ and its transmission time, denoted $T_t^{\text {transmission}_k}$, formulated in Equation~\ref{eq:latency}. 

\begin{equation}
l_t^k = T_t^{\text {local}_k}+T_t^{\text {transmission}_k}
\label{eq:latency}
\end{equation}

\subsubsection{Local computing time}
It corresponds to the computation time required for a local iteration of the gradient descent method. To estimate this time, we rely on the available computing resources of each client $k$, specifically its CPU frequency $f_{t,k}$ during training round $t$, the number of cores in its processor $c_k$ and the number of CPU cycles required to train one bit of data $g_k$. The formula for calculating the local training time of a client $k$ at training round $t$ is described in Equation~\ref{eq:localtime}.

\begin{equation}
\label{eq:localtime}
T_t^{\text {local}_k} = \frac{|D_{t,k}| g_k}{c_k f_{t,k}}
\end{equation}

\noindent where $|D_{t,k}|$ represents the size of the local data in bits held by the client $k$ at training round $t$.

\subsubsection{Transmission time}
It represents the duration needed to transfer the local model parameters from the client to the server. To compute this time, we use the bandwidth of client $k$ at training round $t$, denoted $b_{t,k}$. The formula for calculating the transmission time of client $k$ at training round $t$ is defined in Equation~\ref{eq:transmissiontime}.

\begin{equation}
\label{eq:transmissiontime}
T_t^{\text {transmission}_k} = \frac{\sigma(m_k)}{b_{t,k}}
\end{equation}

\noindent where $\sigma(m_k)$ represents the size of the local model $m$ maintained by the client $k$.

\subsection{Handling Static Heterogeneity: Reputation and Utility}
\label{sec:Staticheterogneity}
To mitigate the effects of static heterogeneity in FL, it is crucial to establish a scoring mechanism to evaluate the contributions of participating clients during each training round. 

\subsubsection{Reputation-based function}
More recently, researchers have extensively explored the adoption of reputation in client selection systems within the realm of FL~\cite{wang_novel_2020}. 

\smallskip
\noindent 
We formulate a reputation-based function as in Equation~\ref{eq:reputationfunction1}. In which, $A_t^k$ represents the accuracy of user $k$'s local model at training round $t$, while $A_{t-1}$ denotes the accuracy of the global model at the previous training round. This equation, denoted as  $\Psi$, is formulated as a recursive function that considers the client's historical contributions.

\noindent Our function addresses the limitation of RFFL~\cite{wang_novel_2020} by including the client's past performance in reputation calculation, avoiding potential issues. This approach considers past contributions for a fairer and more comprehensive evaluation of client performance, overcoming the problem of oscillations in the model's performance.

\smallskip

\begin{equation}
\label{eq:reputationfunction1}
\Psi_t^k=\lambda\left(A_t^k-A_{t-1}\right)+(1-\lambda) { \Psi }{ }_{t-1}^k
\end{equation}

\noindent 
The equation consists of two terms. The first term prioritizes clients whose local models achieve higher accuracy than the previous global model. The second term considers a client's past contributions, reflecting its historical performance compared to the old global model. The hyperparameter $\lambda$ adjusts the balance between current and past contributions.


\smallskip
\noindent 
However, it is important to acknowledge that a significant drawback of this approach is the extensive runtime required for calculating the accuracy of each selected client during every training round according to the validation set. To address this challenge, we introduce a function called the \textit{utility function}, which allows us to overcome this computational burden. 


\subsubsection{Utility function}
The utility function denoted $\zeta$, serves as a measure of the relevance of a client's local update to the global model. Similar to the reputation-based function, the utility function has also been employed for the selection of the most pertinent clients to participate in the training round~\cite{chenning_PyramidFL_2023}. Our utility function is based on normalized model divergence~\cite{fu_client_2022} which quantifies the average difference between the model weights of client $k$ and the global model during training round $t$, and is defined in Equation \ref{eq:utilityfunction1}.

\begin{equation}
\label{eq:utilityfunction1}
d(w_t^k, w_t)= \frac{1}{|w_t|} \sum_{j=1}^{|w_t|}\left|\frac{w_t^{k, j}-w_t^j}{w_t^j}\right|
\end{equation}

\noindent where $w_t$ represents the weights of a model during training round $t$, and $w_t^{k, j}$ and $w_t^j$ is the $j$-th model weight of client $k$ and the global model at training round $t$, respectively.

\smallskip
\noindent This equation assumes that if the normalized model divergence is small, the local update from a client is considered insignificant, resulting in a small value. 
However, this assumption does not always hold true. In cases where the performance of our global model has improved, it implies that clients who have contributed the most to this convergence, i.e., having a minimum distance from the global weight, are actually more significant and should be assigned a higher utility value. This intuition stems from the aggregation process, where the local model parameters of selected clients are combined through a weighted sum. Clients with the smallest distance between their local and global model weights following the aggregation process are the ones who have made the most significant contributions during that specific training round. These clients have exerted the greatest influence on the performance of the global model. In this manner, we define our utility function in Equation \ref{eq:utilityfunction2}. $P$ represents the performance measure used in our study, such as accuracy and F1 score.

\begin{equation}
\label{eq:utilityfunction2}
\zeta_t^k =
\begin{cases}
    e^{-|d(w_t^k, w_t)|}, & \text{if }P_t > P_{t-1}  \\
    1 - e^{-|d(w_t^k, w_t)|}, & \text{if } P_t \leq P_{t-1}
\end{cases}
\end{equation}


\noindent 

\begin{figure}[hbt!]
  \centering
  \includegraphics[width=0.45\textwidth]
  {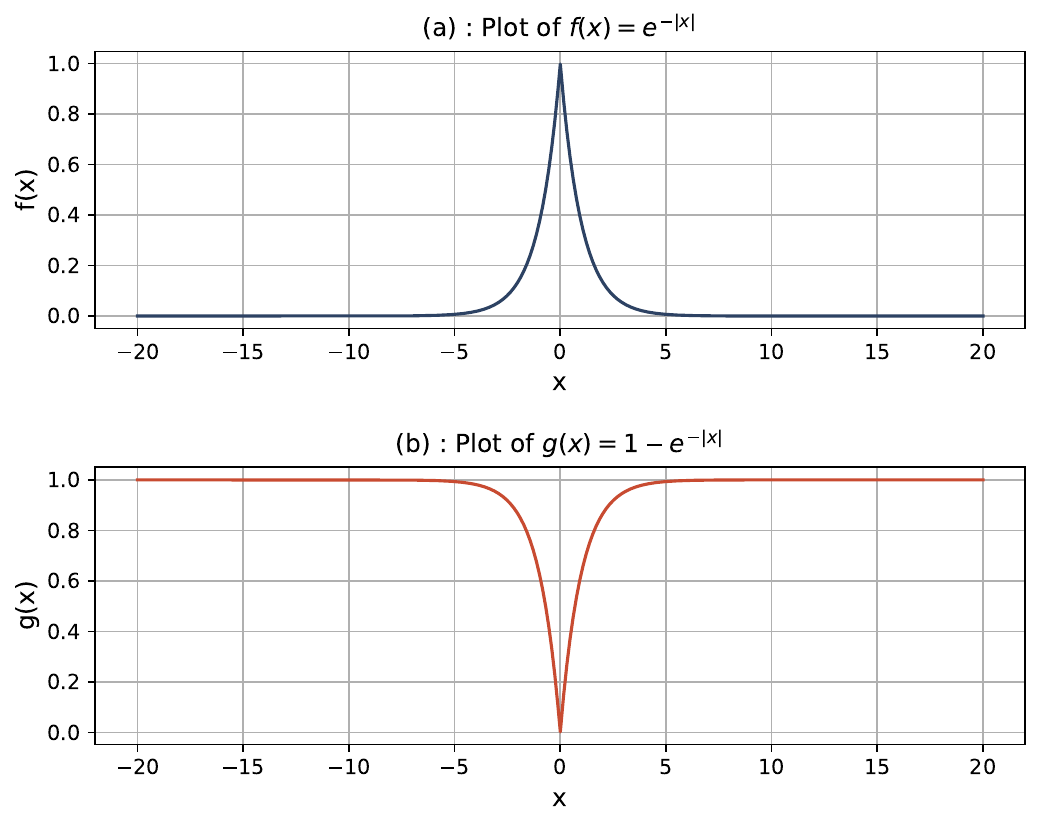}
  \caption{Plot of the two mathematical functions used in our proposed utility function.}
  \label{fig:utility_function}
  \vspace{-0.7cm}
\end{figure}

\smallskip
\noindent Our utility function is defined in two cases. The first case corresponds to an improvement in the global model performance, in which we aim to give a high score to clients whose distances between the parameters of their local model and those of the global model are minimal. To achieve this, we use the function $f(x) = e^{-|x|}$, where $x = d(w_t^k, w_t)$. This function assigns a score close to $1$ for values of $x$ that are near $0$, indicating a high utility. 
Figure \ref{fig:utility_function} illustrates the shape of this function.

\smallskip
\noindent The second case represents a deterioration in the global model performance, in which we intend to assign a low score to clients whose distances between the parameters of their local model and those of the global model are minimal. 
To accomplish this, we utilize the function $g(x) = 1 - e^{-|x|}$, where $x = d(w_t^k, w_t)$. This function assigns a score close to $0$ for values of $x$ that are in proximity to $0$. Figure \ref{fig:utility_function} visually depicts the shape of this function.

\subsubsection{Reputation-based utility function}
To capitalize on the advantages offered by both the reputation and the utility functions, we present a hybridization approach, which is our complete client scoring function. We define this function in Equation \ref{eq:reputility1}. 

\begin{equation}
\label{eq:reputility1}
\Psi_t^k=\lambda \zeta_t^k +(1-\lambda){ \Psi }{ }_{t-1}^k
\end{equation}

\noindent To account for both static and system heterogeneity, we propose a reformulation of the reputation-based utility function described in \ref{eq:reputationutilitylatency}.

\begin{equation}
\label{eq:reputationutilitylatency}
\Psi_t^k=\lambda (\alpha_1\zeta_t^k - \alpha_2l_t^k) +(1-\lambda){ \Psi }{ }_{t-1}^k
\end{equation}

\noindent where $l_t^k$ represents the latency of client $k$ at training round $t$. The inclusion of a negative sign in front of latency in this function serves to guide our RL agent to the minimization of latency. 
$\alpha_1$ and $\alpha_2$ respectively denote the importance assigned to performance maximization and latency minimization.To ensure a mathematically meaningful relationship, we apply Min-Max normalization to the latency value, which allows us to scale it within a range of $[0,1]$.

\smallskip

\subsection{DDQL based client selection}
\label{sec:Methodology}
\begin{figure*}[t]
  \centering
  \includegraphics[width=18cm]
  {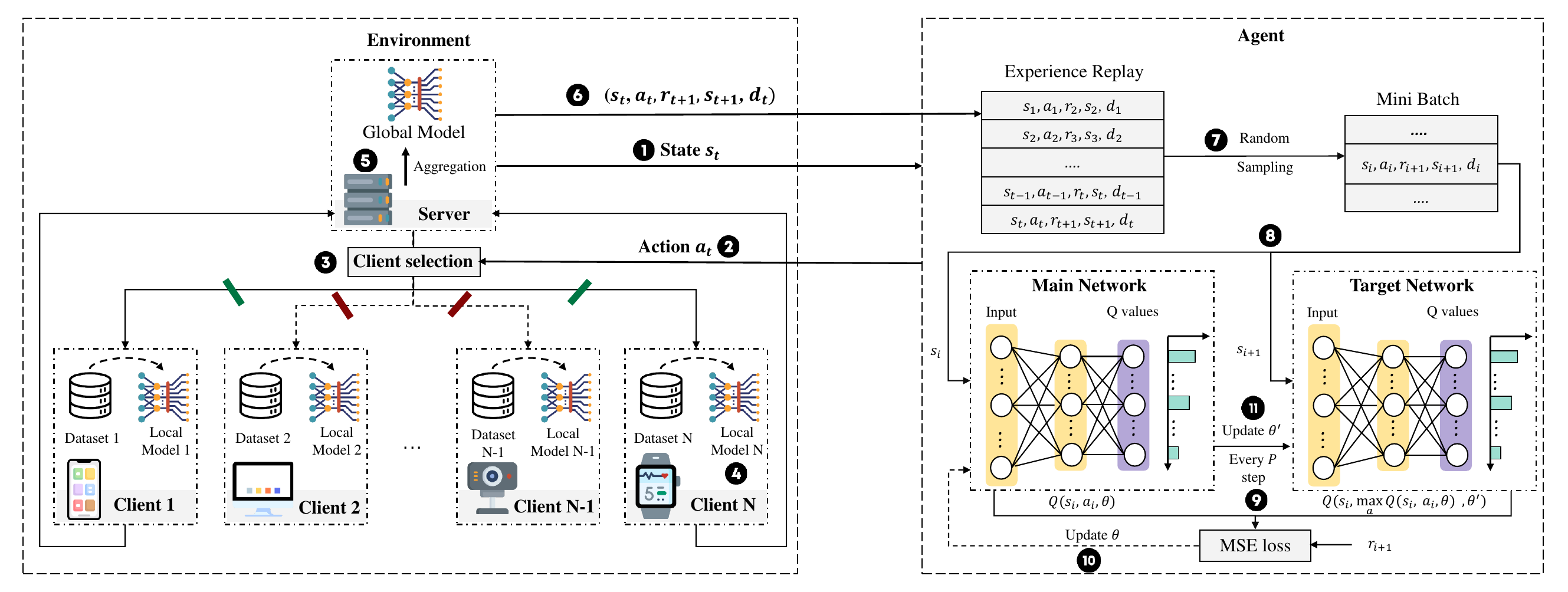}
  \caption{Overview of FLASH-RL methodology. 1- The server sends its current state to the RL agent, 2- The agent sends an action vector to the environment, 3- Client selection based on the action vector, 4- Client local training, 5- Local parameters aggregation, 6- Storage of the current transisition, 7- Random sampling from the replay memory to train our DDRL model, 8- The agent respectively sends the current state and the next state to the main and target networks, 9-The agent calculates the loss using the Q values received from the two networks and the reward, 10-The agent updates the main network parameters using the calculated loss value, 11-If $P$ steps have been completed, the agent transfers the local parameters of the main network to the target network.}
  \label{fig:Global_solution}
  \vspace{-0.5cm}
\end{figure*}

The overall FLASH-RL process is described in detail in Figure \ref{fig:Global_solution}. Details of each step are described below. 

\subsubsection{Markov decision process formulation}
To apply RL to a particular process, it's crucial to formalize it as a Markov decision process. Our formulation includes the key elements: state space, action space, and reward function.

\begin{enumerate}
\item \textbf{State}: 
We introduce a vector representation of the state at each training round $t$, described in  Equation \ref{eq:staterepresentation}.

\begin{equation}
\label{eq:staterepresentation}
s_t=\left(s_t^1, s_t^2, \ldots, s_t^N\right)
\end{equation}

where $s_k$ represents the state of client $k$ at training round $t$ and is defined in Equation \ref{eq:localstate}.

\begin{equation}
\label{eq:localstate}
s_t^k=\left(w_t^k, n_t^k, c_k, f_{t, k}, b_{t, k}\right)
\end{equation}

where $w_t^k,  n_t^k, c_k, f_{t, k},b_{t, k}$ represent respectively the model weights of client $k$, the amount of data held by client $k$, the number of cores of client $k$'s processor, the CPU frequency and the bandwidth at training round $t$.
\smallskip
However, using each client's weight vector directly to train our RL agent's model would be impractical due to its large size. Therefore, we apply the PCA technique to reduce the dimensionality of the weight vector~\cite{wang_optimizing_2020}. 


\smallskip
\item \textbf{Action}: The action is represented by a vector of $N$ booleans, where each component corresponds to the selection status of a specific client. If the $k$-th component equals $1$, client $k$ has been selected. 
This representation is detailed in Equation \ref{eq:action}.

\begin{equation}
\label{eq:action}
a_t=\{i\} \times N, \text { where } i \in\{0,1\}
\end{equation}

However, if we have to select a subset of size $U$ at each training round, adopting the above representation would result in a number of combinations of size $C_N^U$. Such a large action space would raise computational problems and could adversely affect the performance of our RL agent. To solve this problem, we use multi-action RL explained in Section~\ref{sec:DDQLadap}.
\smallskip
\item \textbf{Reward}: The reward is represented by a vector of length $U$, representing the number of clients selected in each training round.

\begin{equation}
r_t=\left(r_t^1, r_t^2, \ldots, r_t^U\right)
\end{equation}

Every element within this vector, $r_t^k$,  corresponds to the reward assigned to client $k$ during training round $t$. Thus, $r_t^k$ equals $\Psi_t^k$.


\end{enumerate}





\subsubsection{Double Deep Q-learning Adaptation}
\label{sec:DDQLadap}
To solve the computational and memory problems associated with using all possible subsets of clients as actions, we propose a modification to the RL algorithm. Our adapted RL approach enables the selection and learning of several actions simultaneously, a concept known as multi-action selection in RL. Instead of considering all possible subsets, we generate a space of $N$ actions representing all clients. Subsequently, we introduce $Q(s_t, a_k)$, which represents the selection value of client $k$ in state $s_t$. To select the $U$ best clients, we consider the $U$ actions with the highest Q value. Initially, we encourage exploration using $\epsilon$-greedy and gradually shift to exploitation by reducing the $\epsilon$ value over training rounds.

\noindent In this work, we used the DDQL algorithm~\cite{latif_survey_2022}. 
Our DDQL agent consists of two neural networks: the main and target network. The main network is used for training, while the target network evaluates the actions of the next state and is updated periodically every $P$ step. Additionally, our DDQL agent incorporates a replay memory mechanism to break the correlation between consecutive experiences $(s_i, a_i, s_{i+1}, r_{i+1}, d_i)$ and $(s_{i+1}, a_{i+1}, s_{i+2}, r_{i+2}, d_{i+1})$, where $d_i$ represents a Boolean value to indicate whether or not the final state has been reached. 

\noindent The RL learning problem can be framed as minimizing the Mean Squared Error loss (MSE) between the target and the approximated values, which is expressed as in Equation \ref{eq:ddqlloss}.

\begin{equation}
\label{eq:ddqlloss}
L_t^k\left(\boldsymbol{\theta}_t\right)=\left(Y_t^k-Q\left(\boldsymbol{s}_t, a_k ; \boldsymbol{\theta}_t\right)\right)^2
\end{equation}

\noindent where $Y_t^k$ t is the target value at round $t$ for action $a_k$ and is defined in Equation \ref{eq:targetvalue1}.

\begin{equation}
\label{eq:targetvalue1}
Y_t^k = r_t^k+\gamma Q\left(\boldsymbol{s}_t, \operatorname{argmax} Q\left(\boldsymbol{s}_t, a_i ; \boldsymbol{\theta}_t\right) ; \boldsymbol{\theta}^{\prime}{ }_t\right)
\end{equation}

\noindent where $r_t^k$ is the reward associated with the selection of client $k$, $\theta$ and ${\theta}^{\prime}$ denotes respectively the parameters of the main and target network, and $\gamma$, represents the discount rate, with $0 \leq \gamma \leq 1$, indicating the value assigned to future rewards.

\noindent This loss is computed for each selected client during training round $t$. 
Rather than formulating the problem as the traditional RL problem of finding the optimal action for a given state, our focus is on identifying the top $U$ actions that correspond to the most suitable clients. 

\section{Evaluation Methodology and Results}
\label{sec:Experiments}
In this section, we present the evaluation of our proposed approach using three benchmark datasets: MNIST, CIFAR-10~\cite{Krizhevsky2009LearningML}, and MobiACT~\cite{Vavoulas2016TheMD} for Fall Detection. 
We compare our approach with existing methods, namely  FedAVG~\cite{mcmahan_communication-efficient_2017}, and FAVOR~\cite{wang_optimizing_2020} to assess its effectiveness and performance.

\subsection{Experimental Methodology}

\paragraph{Dataset Description}
In these experiments, we consider three datasets:

\begin{itemize} 
\item[•] MNIST:  a collection of handwritten digits used for image classification tasks. It includes 60,000 training images and 10,000 test imagess. We train a CNN model with two 5x5 convolution layers. The first layer has 20 output channels, and the second layer has 50. Both layers are followed by 2x2 max pooling.
\item[•]  CIFAR-10: consists of 60,000 colored images divided into 10 classes, with 6,000 images per class. 
We train a CNN model with two 5x5 convolution layers. The first and second layer has 6 and 16 channels respectively. Both layers are followed by 2x2 max pooling.
\item[•]  MobiAct: designed for activity recognition and fall detection. It includes smartphone data collected during various activities, including different types of falls. The dataset comprises 4 types of falls and 12 activities of daily life (ADL). It involves data from 67 volunteers with over 3,200 tests conducted. The dataset captures heterogeneity in terms of gender, age, height, weight, and physical status, contributing to overall heterogeneity bias. For this dataset, we train a two-layer Gated Recurrent Unit model, with each layer consisting of 256 hidden units.

\end{itemize} 

\paragraph{Non-iid data division}
To address the issue of heterogeneity in FL, we focused on non-iid data, which is known to be one of its primary sources. We carefully partitioned our datasets to create subsets that exhibit non-iid distribution.
\smallskip

\noindent To divide MNIST and CIFAR-10 datasets, we used the FedLab framework~\cite{JMLR:v24:22-0440}, which offers various types of data partitioning strategies. For MNIST, we employed the \textit{noniid-label} partitioning, assigning each client data from only two labels with a variable number of data items. Additionally, we used the \textit{Hetero Dirichlet partition} with $\alpha$ values of $0.5$ and $0.8$. 
For CIFAR-10, we adopted the \textit{shards} partitioning, which assigns each client data associated with only two labels, with an equal number of data samples per client. Similarly, we employed the \textit{Hetero Dirichlet partition} with $\alpha$ values of $0.5$ and $0.8$.

\smallskip
\noindent Regarding the MobiAct dataset, it poses some challenges in terms of non-iid data divisions due to its nature as time series data. 
We propose three divisions that will serve as test benchmarks for our solution. In all of these divisions, we have extracted 17 volunteers to form our test set, while the remaining volunteers will be divided among the clients.

\begin{itemize} 
\item[•] \textbf{Volunteer-based division:} Each volunteer in the MobiAct dataset is treated as an individual client residing in their respective smart home, resulting in inherent non-iid division due to their unique behaviors and characteristics.
\item[•] \textbf{Label skew 8:} Each volunteer is treated as a separate client residing in their own smart home. 
They are assigned 8 activities that differ from those assigned to other clients. This introduces another form of non-iid data distribution known as \textit{label skew}.
\item[•] \textbf{Label skew 4:} Similar to label skew 8, but each client is assigned exactly 4 activities instead of 8. 
\end{itemize}

\paragraph{Heterogeneous Hardware settings}
To create a heterogeneous FL environment in terms of system capacity, we introduced variations in the edge equipment and transfer protocols, which are described in table \ref{tab:edgeeqipement} and \ref{tab:transferspec}. These values are set to simulate various edge platforms inspired by: Intel Core i3, i5, i7, first-generation TPU, raspberry PI3, Jetson Nano, Jetson TX2, Jetson Xavier NX, and FPGA Xilinx ZCU102.
This allowed us to mimic the diversity of resources and communication capabilities present in real-world scenarios.

\begin{table}[ht]
\centering
\caption{Edge equipment simulated specifications.}
\vspace{-0.3cm}
\begin{tabular}[t]{lcc}
\hline
Name&CPU Frequency (MHZ)& Number of Cores\\
\hline
Hardware Spec. 1 & 921&128\\
Hardware Spec. 2&1300&256\\
Hardware Spec. 3&800&384\\
Hardware Spec. 4&1100&384\\
Hardware Spec. 5&1377&384\\
Hardware Spec. 6&350&4\\
Hardware Spec. 7&1500&4\\
Hardware Spec. 8&700&1\\
Hardware Spec. 9&3950&2\\
Hardware Spec. 10&4300&4\\
Hardware Spec. 11&4400&4\\
Hardware Spec. 12&4400&8\\
\hline
\label{tab:edgeeqipement}
\end{tabular}
\vspace{-0.8cm}
\end{table}%

\begin{table}[ht]
\centering
\caption{Transfer protocols specifications.}
\vspace{-0.3cm}
\begin{tabular}[t]{lc}
\hline
Name&Bandwidth (Mb/s)\\
\hline
Wi-Fi 1&6\\
Wi-Fi 3&33\\
Wi-Fi 4&336\\
Fast Ethernet&100\\ 

\hline
\vspace{-0.5cm}
\label{tab:transferspec}
\end{tabular}
\end{table}%

\noindent The frequency and bandwidth values specified in the tables represent the average between the maximum and minimum values. To simulate a realistic scenario, with diverse workloads, we modeled each value as a normal distribution with a variable standard deviation. In each training round, a random value is sampled from the respective normal distribution.

\paragraph{Hypeparameter calibration}
 An overview of the hyperparameter values used in our study for the three datasets (Table \ref{tab:hyperparametere}). 

\begin{table}[ht]
\centering
\caption{Hypeparameter tuning.}
\begin{tabular}[t]{lccc}
\hline
Hyperparameter&MobiACT&CIFAR10&MNST\\
\hline
N (number of clients)&50&100&100\\
U (size of selected clients)&5&10&10\\
E (number of local epochs)&5&5&5\\
B (size of batch size)&50&10&50\\
Learning rate&0.001&0.01&0.01\\
Momentum&0.99&0.9&0.9\\
RL batch size&50&50&50\\
$P$&10&10&10\\
RL learning rate&0.01&0.01&0.01\\
$\gamma$&0.9&0.9&0.9\\
$\epsilon_{init}$&0.9&0.9&0.9\\
$\epsilon_{end}$&0.2&0.2&0.35\\
$\Psi_{init}$&$1/50$&$1/100$&$1/100$\\
$\lambda$&0.6&0.6&0.6\\
\hline
\label{tab:hyperparametere}
\vspace{-0.5cm}
\end{tabular}
\end{table}%


\subsection{Overall Results}
Table \ref{tab:overallresults} showcases multiple tests conducted on the non-iid divisions of CIFAR-10 and MNIST datasets. The evaluation metrics employed include accuracy(\%) and latency(s).

\begin{table*}[ht!]
\centering
\caption{Overall results on several dataset divisions compared to FedAvg~\cite{mcmahan_communication-efficient_2017} and FAVOR~\cite{wang_optimizing_2020}.}
\begin{tabular}[t]{ccccccccc}
\hline
Dataset&Division&\#Training rounds&\multicolumn{2}{c}{FeAvg~\cite{mcmahan_communication-efficient_2017}}&\multicolumn{2}{c}{FAVOR~\cite{wang_optimizing_2020}}&\multicolumn{2}{c}{FLASH-RL} \\
&&&Accuracy&Latency&Accuracy&Latency&Accuracy&Latency\\

\hline
&Hetero Dirchlet 0.8&300&52.24&383.43&51.83&388.36&\textbf{54.50}&\textbf{380.45} \\
CIFAR-10&Hetero Dirchlet 0.5&300&50.99&383.92&\textbf{52.90}&382.85&52.25&\textbf{288.59} \\
&Shards&300&47.84&378.61&50.34&380.59&\textbf{51.20}&\textbf{337.86}\\
\hline

&Hetero Dirchlet 0.8&100&99.04&676.89&\textbf{99.09}&682.22&99.06&\textbf{601.07}\\
MNIST&Hetero Dirchlet 0.5&100&98.99&674.13&\textbf{99.04}&672.77&98.90&\textbf{621.06} \\
&Non-iid label&150&98.89&1002.57&\textbf{99.02}&1015.77&98.74&\textbf{979.79}\\
\hline
\label{tab:overallresults}
\vspace{-0.5cm}
\end{tabular}
\end{table*}%

\noindent In the CIFAR-10 dataset, specifically in the Hetero Dirichlet 0.8 division, our approach achieved the highest accuracy compared to FedAVG and FAVOR, with an improvement of 2.26\% and 2.67\% respectively. 
Despite achieving this accuracy improvement, FLASH-RL also managed to reduce latency by 0.78\% compared to FedAVG and 2.03\% compared to FAVOR. 
For the Hetero Dirichlet 0.5 division, FLASH-RL showed a 1.26\% improvement in accuracy compared to FedAVG. 
Although it did not surpass FAVOR in terms of accuracy, FLASH-RL significantly reduced latency by 24.83\% and 24.67\% compared to FedAVG and FAVOR respectively. 
In the Shards division of the CIFAR-10 dataset, FLASH-RL showed a 3.36\% improvement in accuracy compared to FedAVG and a 0.86\% improvement compared to FAVOR. 
Additionally, we observed a reduction in latency by 10.76\% and 11.23\% compared to FedAVG and FAVOR respectively. 
For MNIST, FAVOR achieved the best accuracy results with a slight difference compared FLASH-RL. 
However, our approach reduces the latency. For example, in the Hetero Dirichlet 0.8 division, we achieved a latency reduction of 11.20\% compared to FedAVG and 12.41\% compared to FAVOR.

\begin{figure}[t]
  \centering
  \includegraphics[width=0.4\textwidth]
  {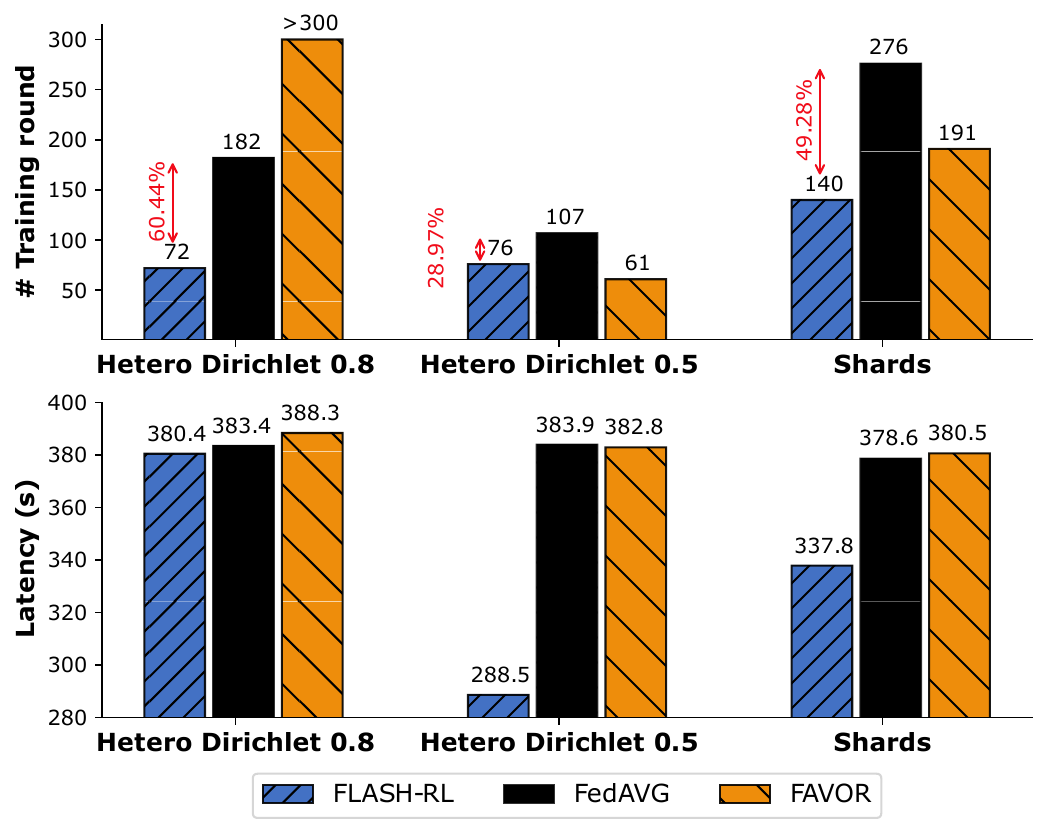}
  \caption{Comparison of training rounds and end-to-end latencies on CIFAR-10.
  }
  \vspace{-0.5cm}
  \label{fig:trainingrounds}
\end{figure}

\noindent This results highlights the effectiveness of our method in striking a desirable balance between maximizing accuracy and minimizing latency. However, an important consideration in FL is the convergence speed of the algorithm, which is typically measured by the number of training rounds required to achieve specific target performance. To assess the convergence speed of our algorithm, we set target accuracies of 52\% for CIFAR-10 Hetero Dirichlet partition 0.8, 50.5\% for CIFAR-10 Hetero Dirichlet partition 0.5, and 47\% for CIFAR-10 shards partition, and we showcase the results in Figure \ref{fig:trainingrounds}.

\noindent 
In the Hetero Dirichlet 0.8 division, FLASH-RL achieved a significant reduction of 60.44\% in the number of training rounds compared to FedAVG, and a remarkable reduction of +76\% compared to FAVOR. In the Hetero Dirichlet 0.5 division, we reduced training rounds by 28.97\% compared to FedAVG. We don't surpass FAVOR in this division, but we have considerably reduced the end-to-end latency, as shown in the second graph. In the shards division, FLASH-RL achieved a reduction of 49.28\% in training rounds compared to FedAVG and 26.70\% compared to FAVOR.

\subsection{Case study}

We now present the results obtained on the divisions of the MobiAct dataset. These results provide insights into the performance and efficacy of FLASH-RL in addressing fall detection within a smart home environment. The evaluation metrics employed include F1-score(\%) and latency(s). Since FAVOR is not specifically designed for fall detection in FL and there is no existing work on applying client selection in this context, we use FedAVG as a baseline for comparison.



\begin{figure}[t]
  \centering
  \includegraphics[width=0.5\textwidth]
  {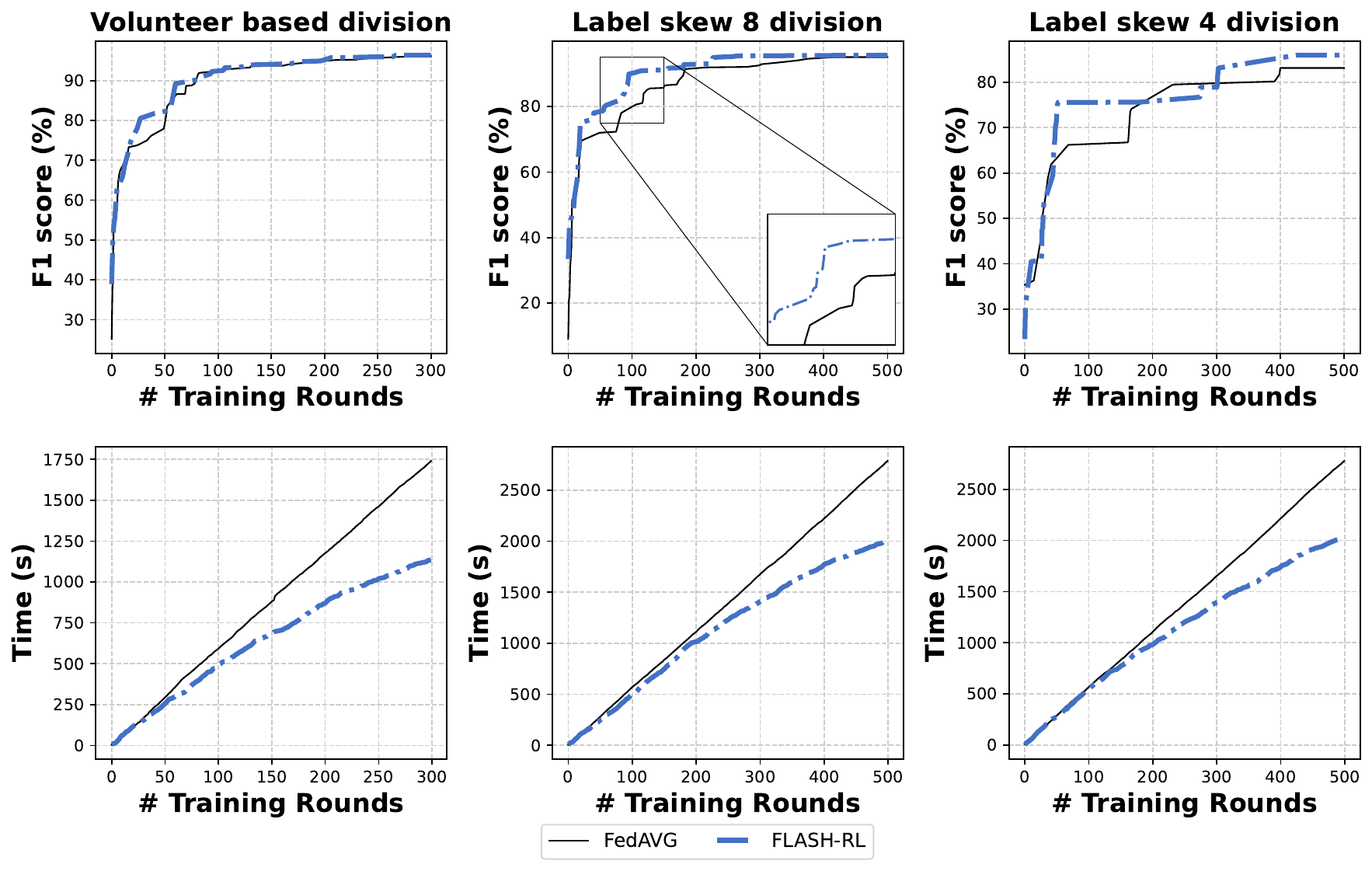}
  \caption{General evolution of FL processes on MobiAct.}
  \label{fig:overallevolution}
  \vspace{-0.5cm}
\end{figure}

\smallskip
\noindent To evaluate the convergence speed and the effectiveness of our algorithm in selecting clients based on reputation, we visualize the F1-score and latency across training rounds (see Figure \ref{fig:overallevolution}).

\smallskip

\noindent The figure illustrates the convergence behavior of our algorithm in terms of F1-score and latency. It is evident that FLASH-RL has faster convergence towards the F1-score targets compared to the FedAVG algorithm. For instance, when setting the F1-score target to 95\% for the label skew 8 division, FedAVG requires 417 training rounds to reach it, while FLASH-RL achieves the same target in only 228 training rounds, resulting in a significant reduction of 45.32\% in training rounds. When setting the F1 score target at 83\% for the label skew division 4, the FedAVG algorithm requires 400 training rounds to reach it. In contrast, FLASH-RL achieves the same target in only 86 training rounds, resulting in a substantial reduction of 78.5\% in the number of required training rounds. Additionally, it is noteworthy that the end-to-end latency by the FedAVG algorithm exhibits a linear trend, whereas FLASH-RL converges towards a constant latency. This observation indicates that our algorithm effectively identifies the best clients, striking a balance between maximizing F1-score and minimizing latency.


\section{Conclusion}
\label{sec:Conclusion}
In this study, we have proposed FLASH-RL, a novel FL  framework that leverages DDQL to effectively address the challenges of system and static heterogeneity in FL. 
Our approach addresses the problem of selecting the most suitable clients for each training round, considering their performance, latency, and past contributions. 
By integrating a novel reputation-based utility function into DDQL's reward mechanism, we have evaluated and ranked clients based on their contributions to the FL process. 
Furthermore, we have proposed enhancements to the DDQL algorithm, enabling it to make multiple action choices simultaneously, thereby accelerating the convergence of the agent model. FLASH-RL outperforms state-of-the-art strategies on MNIST, CIFAR-10, and MobiAct divisions, enabling faster convergence. For instance, in CIFAR-10, FLASH-RL reduces latency by up to 24.82\% and 24.67\% compared to FedAVG and FAVOR. It also reduces training rounds by up to 60.44\% compared to FedAVG and +76\% compared to FAVOR. 

\bibliographystyle{IEEEtran}
\bibliography{references}

\end{document}